\documentclass[final]{cvpr}

\usepackage{times}
\usepackage{epsfig}
\usepackage{graphicx}
\usepackage{amsmath}
\usepackage{mathtools, amssymb}
\usepackage{multirow}
\usepackage{amsthm}

\def\confYear{CVPR 2021}
\begin{document}

\title{SBNet: Segmentation-based Network for Natural Language-based Vehicle Search}

\author{Sangrok Lee\\
MODULABS\\
{\tt\small srl@modulabs.ai}
\and
Taekang Woo\\
NAVER Corporation\\
{\tt\small t.k.woo@navercorp.com}

\and
Sang Hun Lee\thanks{Corresponding author}\\
Kookmin University\\
{\tt\small shlee@kookmin.ac.kr}
}


\maketitle

\begin{abstract}
   Natural language-based vehicle retrieval is a task to find a target vehicle within a given image based on a natural language description as a query. This technology can be applied to various areas including police searching for a suspect vehicle. However, it is challenging due to the ambiguity of language descriptions and the difficulty of processing multi-modal data. To tackle this problem, we propose a deep neural network called SBNet that performs natural language-based segmentation for vehicle retrieval. We also propose two task-specific modules to improve performance: a substitution module that helps features from different domains to be embedded in the same space and a future prediction module that learns temporal information. SBnet has been trained using the CityFlow-NL dataset that contains 2,498 tracks of vehicles with three unique natural language descriptions each and tested 530 unique vehicle tracks and their corresponding query sets. SBNet achieved a significant improvement over the baseline in the natural language-based vehicle tracking track in the AI City Challenge 2021. Source Code: \url{https://github.com/lsrock1/nlp_search}
\end{abstract}

\section{Introduction}
Searching for a vehicle in an image database with natural language (NL) descriptions is a challenging problem in computer vision\cite{Feng21CityFlowNL, Li2017PersonSW}. It can be widely applied in video surveillance and traffic analysis. Currently, many surveillance cameras are installed on roads and highways and generate a huge amount of video data per second. Manually searching for a criminal suspect vehicle in such video data can be time consuming. Therefore, the development of an automatic vehicle search function is vital and urgent. 

Vehicle retrieval using image-based queries is called vehicle re-identification in computer vision\cite{Khan2019, Zakria2021, Sangrok2020StRDAN, musp2021}. Given a query image, the algorithm obtains affinities between the query and the vehicle images in the database and retrieves the most similar vehicles. However, in many criminal cases, there are no images of the suspected vehicle, only verbal descriptions. Therefore, this method’s ability to find the vehicle is significantly limited in these cases. 

To overcome this limitation, searching for vehicles with NL descriptions has been proposed. It is not necessary to provide a vehicle photo as in the image-based query method. NL also can precisely describe the details of the vehicle appearance and does not require labelers to go through the entire list of attributes.

\begin{figure}[t]
    \label{g_abstract}
    \begin{center}
        \includegraphics[width=1.0\linewidth]{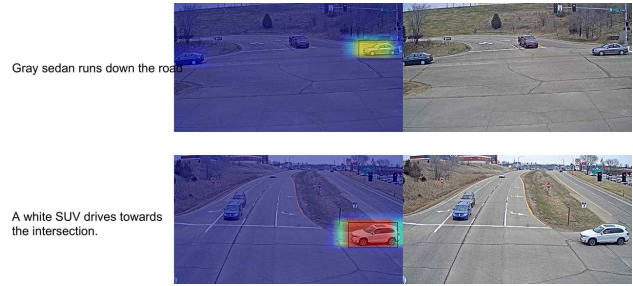}
    \end{center}
    \caption{Examples of proposed SBNet output. With natural language and an image, it finds the corresponding area in the image and shows high activation in that area. The left column is the natural language description, the center column is the model’s output activation mask, and the right column is the input image.}
\end{figure}

In this work, the CityFlow-NL dataset \cite{Feng21CityFlowNL} is used as the benchmark dataset. According to our observations, the main problems in this task are multi-modal question answering and relational reasoning. The task can be assumed as answering a visual question where the answer is yes or no. We adopt the attention mechanism to deal with the multi-modal dataset and the channel modulation method proposed in \cite{Perez2018FiLMVR}. In addition to this module, we propose future prediction and substitution modules to improve performance. The future prediction module is for embedding vehicle movement information, while the substitution module helps to describe two different types of domain data in the same embedding space. Figure \ref{g_abstract} illustrates our model's activation results for given sets of image and NL description. Our contributions are summarized below.

\begin{itemize}
\item We propose a new segmentation-based network model called SBNet to perform NL-based vehicle retrieval.
\item We introduce two specific modules to improve performance: the future prediction and substitution modules.
\item Our proposed SBNet outperforms the current baseline model without post-processing.
\end{itemize}


\section{Related Work}
Natural language-based vehicle retrieval is a multi-target and multi-camera task that uses multi-modal data for images and NL descriptions.
Other tasks like object tracking via descriptions\cite{Feng2020RealtimeVO, Feng2019RobustVO, Li2017TrackingBN}, video retrieval\cite{Zhang2019MANMA, Hendricks2017LocalizingMI}, video localization\cite{Gavrilyuk2018ActorAA, Hu2016SegmentationFN,Yamaguchi2017SpatioTemporalPR} shares some similar points with the NL-based vehicle retrieval. In tracking tasks, they employ detection models and leverage language features from the hidden states of the recurrent neural network(RNN)\cite{Feng2020RealtimeVO}.
Liu \etal \cite{Li2017TrackingBN} used RNN and convolutional neural network(CNN) models to extract embedding features and also utilized the attention mechanism and dynamic filter generation method.
In video retrieval and localization, Zhang \etal\cite{Zhang2019MANMA} introduced a video embedding model and graph representation.
Hendricks \etal \cite{Hendricks2017LocalizingMI} exploited the context network and distance loss to embed different modal data to the same feature space. Gavrilyuk \etal \cite{Gavrilyuk2018ActorAA} employed CNN architectures for the language and video backbone model to predict the segmentation mask of the target object.

NL-based person re-identification (re-id) is also a very similar task with NL-based vehicle one \cite{Li2017PersonSW} except for different targets just as personal re-id is very close to vehicle re-id. However, in NL-based person re-id, only cropped person images are given, and thus understanding the background context is not required unlike in NL-based vehicle re-id. The spatial-temporal localization by NL description task was first introduced by Yamaguchi \etal\cite{Yamaguchi2017SpatioTemporalPR, Feng21CityFlowNL}. The ActivityNet dataset is annotated with NL descriptions to facilitate the training and evaluation of the proposed task. However, the temporal retrieval in \cite{Yamaguchi2017SpatioTemporalPR} entails retrieving the target video clip from a set of video clips. On the contrary, the goal of the vehicle retrieval task is to temporally localize the target object within one sequence of video. Additionally, the targets in the ActivityNet-NL take up most of the frame and cannot serve as a tracking benchmark.

Feng \etal\cite{Feng21CityFlowNL} extended the widely adopted CityFlow Benchmark with NL descriptions for vehicle targets and introduce the CityFlow-NL Benchmark. The CityFlow-NL contains more than 5,000 unique and precise NL descriptions of vehicle targets, making it the first multi-target multi-camera tracking with NL descriptions dataset. Moreover, the dataset facilitates research at the intersection of multi-object tracking, retrieval by NL descriptions, and temporal localization of events. They focus on two foundational tasks: the Vehicle Retrieval by NL task and the Vehicle Tracking by NL task, which take advantage of the proposed CityFlow-NL benchmark and provide a strong basis for future re- search on the multi-target multi-camera tracking by NL description task.

We introduced the CityFlow-NL Benchmark to develop SBNet, an NL-based vehicle retrieval network. Our approach proposed in this paper is close to image localization methods. However, we also consider the temporal information using the future prediction module and employ the co-attention mechanism and channel modulation to embed relation features. 

\begin{figure*}
    \label{architecture}
    \begin{center}
        \includegraphics[width=0.95\linewidth]{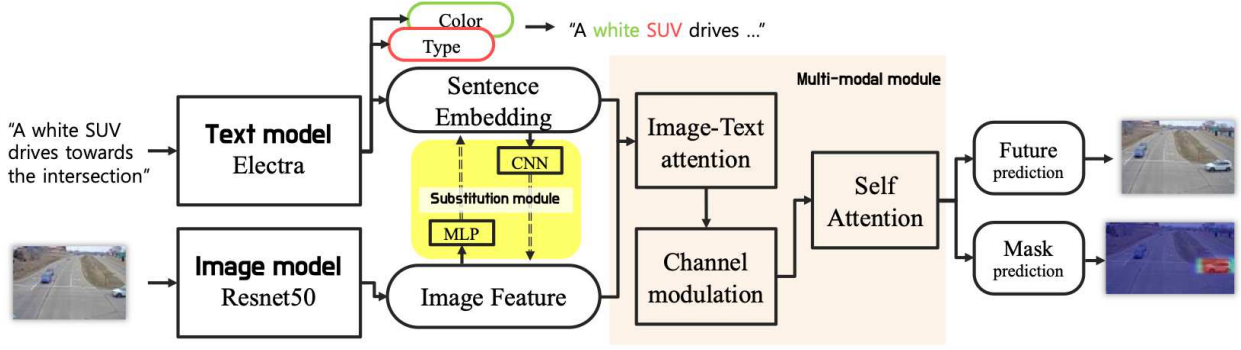}
    \end{center}
    \caption{Overall architecture of SBNet.}
\end{figure*}

\section{Problem Definition}

In the AI City Challenge, the goal of the NL-based vehicle retrieval task is to find a vehicle described in English from images. Vehicles are presented as objects with bounding boxes for each image in the scene video, not as cropped objects. The image can have multiple vehicles, and the target vehicle is specified by a bounding box. Three different NL descriptions are provided for one target vehicle. 

The goal of this task is to segment the area where the corresponding vehicle is located when the query is given. For this task, we prepared a train dataset $D_{train}$ and a set of the ground truth segmentation labels $G$:
\begin{multline}
D_{train} = \{(i_1, n_1^1, n_1^2, n_1^3), 
    (i_2, n_2^1, n_2^2, n_2^3), 
    \dots , \\ 
    (i_t, n_t^1, n_t^2, n_t^3)\},
\end{multline}
\begin{equation}
    G = \{(g_1, g_2, g_3, \dots , g_t), \}, g \in R^{h \times w},
\end{equation}
where $D_{train}$ has $t$ sets of data, each of which has image($i_j$), and three NL description($n_j^k$). $G$ is ground truth segmentation labels for each train data. We introduce bounding box segmentation, which assigns a class to every pixel in the region of bounding box of the target in a given image, and apply it throughout the work. 

\section{Proposed Method}
To tackle the NL-based vehicle retrieval task, we propose a multi-modal localization model. In the following sections, we first illustrate our overall architecture in Section \ref{sec:overall}. Then, we describe the NL module (NLM) and image processing module (IPM) in Section \ref{sec:natural} and Section \ref{sec:image}, respectively. Finally, we introduce the multi-modal module, which handles the NL feature and image feature simultaneously and searches for the target vehicle using the language feature, in Section \ref{sec:multimodal}. 

\subsection{Overall architecture}
\label{sec:overall}

As illustrated in Fig. \ref{architecture}, our proposed method has three main feature modules: the NLM, IPM, and multi-modal module. 
First, the raw image and NL descriptions are embeded using the IPM and NLM. We adopt ResNet50\cite{He2016DeepRL} for the IPM and ELECTRA\cite{Clark2020ELECTRAPT} for the NLM. ResNet50 is a widely used backbone network for image processing, while ELECTRA is a well-known pretrained model for NL. In general, the pretrained NL models, like BERT and ELECTRA, outperform the previous ones and show reasonable performance on subtasks that have few datasets. In our task, we adopt ELECTRA because ELECTRA outperforms the transformer models that are not pretrained\cite{Vaswani2017AttentionIA}. 
In addition, the multi-modal module is one of the most important parts of our network, which was inspired by relational reasoning\cite{Perez2018FiLMVR, Santoro2017ASN}. To combine two types of information, image and NL, co-attention and channel mixture are used. 

The three modules, IPM, NLM, and multi-modal module, are the backbone of our network. The other modules such as future prediction, substitution, and classification modules were introduced additionally for boosting performance. The details are described in the following sections.

\subsection{NL module}
\label{sec:natural}
The description type is English NL. To embed this data, we use the pretrained language model ELECTRA\cite{Clark2020ELECTRAPT}. ELECTRA consists of transformer modules and a self-supervised language representation learning model and has two parts: a generator and a discriminator. In the self-supervised phase, the masked language is fed to the model. The generator predicts adequate words for the mask and the discriminator distinguishes the generated words. In our model, we use ELECTRA’s small discriminator as the NLM. The NLM process is as follows:
\begin{equation}
    FN = NLM(n), FN \in R^{l \times e},
\end{equation}
where $FN$ is the embedded language feature, $l$ is the length of the sentence, and $e$ is the embedding dimension size. In our work, we set $l$ to 30 in training and $e$ to 2048. It is note that ELECTRA small output channel is 256, therefore we use another $256 \times 2048$ linear layer.

\subsection{Image processing module}
\label{sec:image}
We exploit ResNet50 as IPM to embed the scene image.
The image feature is leveraged for segmenting the vehicle box area.
To maintain resolution, we set the stride of the last five stages of ResNet to 1 rather than 2.
\begin{equation}
    FI = IPM(i), FI \in R^{c \times h^* \times w^*},
\end{equation}
where $FI$ is an image feature, $c$ is the embedding channel, and $h^*$ and $w^*$ are feature height and width respectively. $c$ is 2048 in our work.

\subsection{Classification module}
In addition to the NLM and IPM, we attach simple classification modules to each module.
With a rule-based algorithm, we can extract the vehicle color and type.
Using these labels, we can attach classification modules to NLM and IPM modules.
In ELECTRA, $CLS$ token is used for the classification task.
Following this, we use the $CLS$ token position to classify color and type in the NLM.
In the IPM, we pool a vector via the bounding box and classify the color and type with this vector.
\begin{equation}
    C_n, T_n = L_{n}(FN_{cls}),
\end{equation}
\begin{equation}
    FI_{cls} = \frac{1}{|b|}\sum_{i=0}^{h^*}\sum_{j=0}^{w^*}{FI * b}, b \in R^{1 \times h^* \times w^*}, FI_{cls} \in R^{c},
\end{equation}
\begin{equation}
    C_i, T_i = MLP_i(FI_{cls}),
\end{equation}
where $C_n$ and $T_n$ are the predicted color and type from the NL feature.
$FN_{cls}$ is the $CLS$ token feature and $L_n$ is one linear layer.
$FI_{cls}$ is the masked pooled vector in the spatial dimension.
Via $FI_{cls}$, we predict $C_i$ and $T_i$, which are the color and type of the vehicle in the image.
$MLP_i$ has two linear layers with a ReLU function.

\subsection{Substitution module}
This task assumes a one-to-one relation between each image and language set.
In other words, an image and NL descriptions are semantically same.
Inspired by this idea, we devise and add a substitution module.
This module generates an image feature from an NL feature and an NL feature from an image feature.
In learning, we try to train these two features so that they are exchangeable.
This module's target features that should be generated are as follows:
\begin{equation}
    FI_{gt} = \frac{1}{h^*\times w^*} \sum_{i=0}^{h^*}\sum_{j=0}^{w^*}{FI^{(i)(j)}}, FI_{gt} \in R^{c},
\end{equation}
\begin{equation}
    FN_{gt} = \frac{1}{l} \sum_{k=0}^{l}{FN^{(k)}}, FN_{gt} \in R^{e},
\end{equation}
where $FI_{gt}$ is the target image feature that is the spatial mean of $FI$, 
$FI^{(i)(j)}$ is a vector on the $i, j$ point in the spatial dimension,
$FN_{gt}$ is the target NL feature that is the mean of the sentence feature,
$FN^{(k)}$ is a vector of the $k^{th}$ word feature.
The generated features are summarized as follows:
\begin{equation}
    FI_{g} = MLP(FN), FI_{g} \in R^{c},
\end{equation}
\begin{equation}
    FN_{g} = CNN(FI, b), FN_{g} \in R^{e},
\end{equation}
The $MLP$ is two stacked linear layers with one leaky ReLU activation function. The $CNN$ consists of two convolutional layers with one ReLU activation function, and $b$ is the target vehicle's bounding box.

\subsection{Multi-modal module}
\label{sec:multimodal}
Multi-modal module is for interpreting two different domain features: image and NL.
First, we apply co-attention\cite{Hsieh2019OneShotOD} with the image and NL feature.
Using an attention matrix, we enhance the NL and image feature information from a different domain.
Using channel modulation\cite{Hu2020SqueezeandExcitationN}, we inject the relation information of the query description.
Finally, we exploit the self-attention method for relation information in the spatial domain.
The attention between two features $FN$ and $FI$ is defined as follows:
\begin{equation}
    A = al(FI) \otimes an(FN),
\end{equation}
\begin{equation}
    A = \sigma(\frac{A}{\sqrt{c}}), A \in R^{l \times h^*w^*},
\end{equation}
$A$ is the resulting attention, $al$ is a linear projection layer, and $an$ is a $1 \times 1$ convolutional layer. 
$\otimes$ is matrix multiplication.
$\sigma$ is a softmax function.
We leverage $A$ to enhance the $FN$ and $FI$.
The enhancement is conducted as follows:
\begin{equation}
    FN_e = FN + A \otimes FI,
\end{equation}
\begin{equation}
    FI_e = FI + A \otimes FN
\end{equation}

The language description has relation information in the scene situation such as the following:
\begin{quotation}
\textit {White SUV keeps straight behind a line of vehicles.}
\end{quotation}
Leveraging relation reasoning method we modulate channel activation to inject relation information from language feature.
The modulating process is as follows:
\begin{equation}
    FN_m = FN_e * MLP(FI_e),
\end{equation}
The $MLP$ has two linear layers, with ReLU activation as the intermediate layer and the sigmoid function as the end layer.
The result of $MLP$ is in the $c \times 1 \times 1$ dimension.
We compute element-wise multiplication. 

\subsection{Mask prediction module}
Finally, to conduct main task, vehicle area segmentaion, we adopt multiple convolutional layers to segment the mask.
The segmentation layers are follows:
\begin{equation}
    M = G(FN_m), M \in R^{1 \times h^* \times w^*},
\end{equation}
where $M$ is the mask and $G$ is the convolutional layers (i.e., three non-linear layers and one convolution layer). 
The non-linear layers have convolution, batch normalization, and ReLU functions. 
All convolutional layers have a $3 \times 3$ kernel size and a $1 \times 1 $ padding size.
The output channels are $1024$, $512$, $256$, and $1$ in order.

\subsection{Future prediction module}
The image is part of a video.
Thus, the description involves temporal information such as the following:
\begin{quotation}
    \textit {A gray small car is turning left.}
\end{quotation}
To embed this vehicle action, we add the future frame prediction module.
This simply predicts the next frame using $FI$.
The prediction is conducted as follows:
\begin{equation}
    U = B(FI), U \in R^{3 \times h^* \times w^*},
\end{equation}
where $U$ is the predicted future frame, and $B$ is the multiple convolutional layers.

\subsection{Probability computation}
\begin{figure}[t]
    \label{probability}
    \begin{center}
        \includegraphics[width=0.75\linewidth]{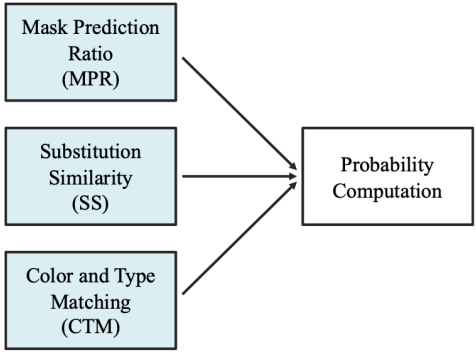}
    \end{center}
    \caption{The probability computation process between the NL description and image.}
\end{figure}

To perform the task, we compute the probability of matching between the NL description and vehicle.
As illustrated in Fig. \ref{probability}, The probability has three parts: mask prediction ratio (MPR), substitution similarity (SS), and color and type matching probability (CTM). 

\begin{equation}
    MPR = \frac{\sum{M * b}}{\sum{b}},
\end{equation}

\begin{equation}
    SS = cs(FI_{gt}, FI_g) + cs(FN_{gt}, FN_g),
\end{equation}
where $cs$ is the cosine similarity, and 

\begin{equation}
    CTM = C_i[C_n] + T_i[T_n],
\end{equation}
where $C_i[C_n]$ is the output of the softmax value of $C_i$ of the $C_n$ color element, and $T_i[T_n]$ is the output of the softmax value of $T_i$ of the $T_n$ type element.

The final probability between a NL description and an image is computed as follows:
\begin{equation}
    Prob = MPR + SS + \lambda * CTM
\end{equation}
where $\lambda$ is a weight and is set to 0.5.

\section{Loss Function}

The total loss function is defined as follows:
\begin{equation}
    L_{total} = L_{seg} + \lambda_1 \times L_{cls} + L_{sub} + \lambda_2 \times L_{fut}
\end{equation}
where $L_{seg}$ is vehicle segmentation loss, $L_{cls}$ is vehicle classification loss, $L_{sub}$ is substitution loss, $L_{fut}$ is future prediction loss, and $\lambda_1$ and $\lambda_2$ are weights and are set to 0.2 and 0.2 respectively. The losses are described in the following sections.

\subsection{Vehicle segmentation loss}
Vehicle segmentation loss is binary mask prediction loss which is calculated using the binary cross entropy formula as follows:
\begin{equation}
L_{seg} = -\sum b \times log (M),
\end{equation}
where $b$ is the ground truth segmentation mask.
It is noteworthy that we use the bounding box area as the segmentation mask because we do not have a precise vehicle mask.

\subsection{Vehicle classification loss}
Vehicle classification loss is cross entropy loss for $C_n$, $T_n$, $C_i$, and $T_i$.
We define the prediction set as $P = \{C_n, C_i, T_n, T_i\}$ and the corresponding ground truth set as $P_{gt}$:
\begin{equation}
    L_{cls} = -\sum_{Z\in P} P_{gt}[Z] \times log (Z),
\end{equation}
where $P_{gt}[Z]$ is the ground truth label of $Z$.

\subsection{Substitution loss}
The goal of this loss function is to train the $FI$ and $FN$ so that they are semantically exchangeable.
\begin{equation}
    L_{sub} = 2 - cs(FI_{gt}, FI_g) + cs(FN_{gt}, FN_g),
\end{equation}
where $cs$ is the cosine similarity. This loss makes the generated features and ground truth features closer.

\subsection{Future prediction loss}
This loss is aimed for training the future prediction module so that it forecasts next frame more precisely.
The mean squared error is calculated for each pixel and summed for all pixels.
\begin{equation}
    L_{fut} = \frac{1}{3\times h \times w}\sum (U - I)^2
\end{equation}

\section{Dataset}

\subsection{Cityflow-NL}
Our model has been developed using the Cityflow-NL dataset. This dataset consists of 666 targets vehicles in 3,028 (single-view) tracks from 40 calibrated cameras, and 5,289 unique NL descriptions. Each track has a sequence of scene images and three NL descriptions typically. The average number of frames a target vehicle is 75.85.

\subsection{Denoising the descriptions}
To depress the noise in the learning process, we perform pre-processing.
Vehicle color and type are ambiguous in perception.
Therefore, in some cases, the three descriptions are not match in terms of color and type.
We use the extraction and voting method to clean the color and type.
First, we choose 12 colors and 10 types that exist in this dataset as labels.
Via a matching algorithm, we can extract the type and color from the sentence.
Because sentences are simple and mostly have only one color and type specifier, it can be easily conducted.
One image has three NL descriptions, and each has a color and type.
To unify the color and type, we choose the most likely type and color.
Finally, we replace the other colors and types with the chosen color and type to denoise the data.

\section{Experiment}

\subsection{Implementation detail}
Preprocessing resizes all images to 384×384 pixels and applies random translation effects. We use the Adam \cite{Kingma2015AdamAM} optimizer with a weight decay of 3e-5 and a momentum of 0.9. The proposed model is trained with a batch size of 64, a training epoch of 10, and an initial learning rate of 0.00003, divided by 10 at 5 and 8 epochs. Label smoothing is also applied to avoid overfitting in classification loss. Training required 24 h on the Cityflow-NL datasets, using 4 NVIDIA RTX 1080 GPU system. The training code was written in PyTorch \cite{Paszke2019PyTorchAI}.

\begin{figure}[t]
    \label{top_five}
    \begin{center}
    \includegraphics[width=1.0\linewidth]{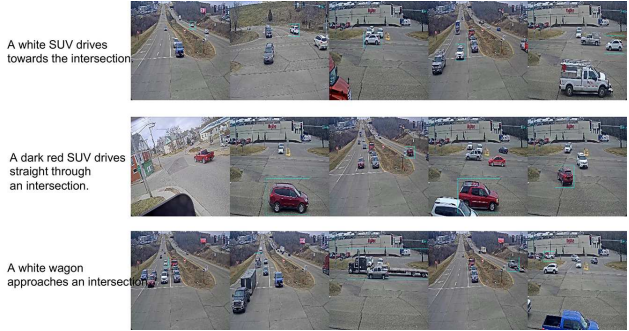}
    \end{center}
    \caption{Examples of five highest matching images for each NL  description. The NL description is on the left, and the images are sorted in descending order according to the matching probability on the right.}
\end{figure}

Figure \ref{top_five} shows the examples of the SBNet's output for five highest matching images for each NL description. NL descriptions are on the left while the images sorted in descending order according to the matching probability are on the right.

\subsection{Evaluation metric and results}
To evaluate, we use recall and Mean Reciprocal Rank (MRR) metric, which are standard metrics for retrieval tasks\cite{Manning2005IntroductionTI}.
We use the baseline model for Cityflow-NL Benchmark presented in \cite{Feng21CityFlowNL}. 

\subsubsection{Effects of modules}

To evaluate the effects of each module, we conduct ablation studies with and without the modules.
Table \ref{modules} shows that the performance gain depends on additional modules.
With each modules, its corresponding probability computation are also added.
The classification module brings about 0.5\% improvement in baseline (i.e., without any additional modules).
The model with the substitution module shows an 1\% improvement, and the future prediction module achieves a 0.7\% performance improvement.

\begin{table}[ht]
    \centering
    \begin{tabular}{ccccc}
        Module Name &&Included&&\\
        \hline
        Classification & No & Yes & Yes & Yes \\
        Substitution & No & No & Yes & Yes\\
        Future Prediction & No & No & No & Yes\\ \hline
        MRR & 0.0977 & 0.1025 & 0.1124 & 0.1195\\
        \hline
    \end{tabular}
    \caption{Model performance on with and without modules from the leaderboard.
    Without the three modules, the model consists of the IPM, NLM, multi-modal module, and mask prediction module.}
    \label {modules}
\end{table}

\subsubsection{Cityflow-NL performance}

Our competition performance is shown in Table \ref{leaderboard}.
We achieve 10th place on the final leaderboard.
We also make a comparison with the baseline proposed in \cite{Feng21CityFlowNL}.
SBNet achieves a significant performance improvement from the baseline as shown in Table \ref{compare_baseline}.

\begin{table}[ht]
    \centering
    \begin{tabular}{ccccc}
        Rank & Name & MRR \\
        \hline
        1 & Alibaba-UTS & 0.1869 \\
        2 & TimeLab & 0.1613\\
        3 & SBUK & 0.1594\\
        4 & SNLP & 0.1571\\
        5 & HUST & 0.1564\\
        6 & HCMUS & 0.1560\\
        7 & VCA & 0.1548\\
        8 & aiem2021 & 0.1364\\
        9 & Enablers & 0.1314\\
        10 & \textbf{Modulabs (ours)} & 0.1195\\
        \hline
    \end{tabular}
    \label{table:1}
    \caption{Leaderboard of the Track 5 in the AI City Challenge 2021.}
    \label{leaderboard}
\end{table}

\begin{table}[ht]
    \centering
    \begin{tabular}{ccccc}
        Model & MRR \\
        \hline
        SBNet & 0.1195 \\
        Siamese baseline\cite{Feng21CityFlowNL} & 0.0269\\
        \hline
    \end{tabular}
    \caption{Performance comparison between the SBNet and baseline models.}
    \label{compare_baseline}
\end{table}

\section{Conclusion}
To tackle the NL-based vehicle retrieval task, we proposed a segmentation-based network model called  SBNet.
It consists of the IPM, NLM, and multi-modal module to handle NL descriptions and images simultaneously.
We also introduce the substitution module and future prediction module, which improve the performance.
The matching probability between the description and image is computed with each module's output.
We achieved a significant improvement over the baseline and ranked 10th in the natural language-based vehicle tracking track in the AI City Challenge 2021.

\section*{Acknowledgement}
This work was supported by the National Research Foundation of Korea (NRF) grant funded by the Korea government (MEST) (No.2020R1A2C1102767).
{\small
\bibliographystyle{ieee_fullname}
\bibliography{SBNet}
}

\end{document}